> Editorial: Advances in Self-Organizing Maps

The Self-Organizing Map (SOM) with its related extensions is the most popular artificial neural algorithm for use in unsupervised learning, clustering, classification and data visualization. Over 5,000 publications have been reported in the open literature, and many commercial projects employ the SOM as a tool for solving hard real-world problems.

Each two years, the "Workshop on Self-Organizing Maps" (WSOM) covers the new developments in the field. The WSOM series of conferences was initiated in 1997 by Prof. Teuvo Kohonen, and has been successfully organized in 1997 and 1999 by the Helsinki University of Technology, in 2001 by the University of Lincolnshire and Humberside, and in 2003 by the Kyushu Institute of Technology. The Université Paris I Panthéon Sorbonne (SAMOS-MATISSE research centre) organized WSOM 2005 in Paris on September 5-8, 2005.

The Paris WSOM conference was very successful, with 150 participants, coming from more than 23 countries from all continents (even from Australia). With the help of nice weather and good wine, the atmosphere was studious as well as friendly thanks to the high level of the communications and to the social program (in particular at the first floor of the Eiffel Tower). The Self-Organized Maps community had once again the opportunity to meet and to share knowledge and expertise.

Since the first publications, the SOM algorithm as defined by Teuvo Kohonen in the seventies has known a large development. Nowadays, it is used as well as other classical methods for a lot of data mining tasks (classification, clustering, reduction of dimension, descriptive statistics, non linear projection, time series analysis, missing data completion, inquiry exploitations, visualization of high dimensional data, etc.). Its mathematical properties have been more deeply studied, even if work remains to be done. A lot of modifications of the original algorithm have been defined in order to allow a more complete theoretical study, or to possess some interesting properties. The range of its applications does not stop to increase, having recently touched the economic and management sciences.

Many years and several thousands of scientific papers after the first publications about self-organized maps, there is still an important activity in the development of new algorithms and methods aimed to build self-organized, topographic maps. In this issue, Kohonen himself proposes a new self-organizing system that can produce superimposed responses to superimposed stimulus patterns. This principle can be used for instance to model pointwise neural projections such as the somatotopic maps. Sullivan et al. use SOM as a model for cortical development. They show that homeostatic synaptic scaling can replace standard weight normalization, as a plausible mechanism for controlling the weight growth associated with Hebbian learning. Van Hulle introduces a new learning algorithm for topographic map formation of Edgeworth-expanded Gaussian activation kernels, and shows the superiority of this approach e.g. in clustering tasks. Ontrup and Ritter introduce a growing variant of the Hyperbolic Self-Organizing Map, combining the advantages of the latter with the power of growing, not a priori defined, lattices. Neural gas methods form an important class of robust clustering methods. Two papers cover Neural Gas: Cottrell, Hammer et al. propose a batch Neural Gas algorithm with faster convergence and applicability to non-vectorial proximity data. Villmann et al. extend Neural Gas for supervised fuzzy classification, including the use of relevance learning. Finally, in the context of kernel methods, Yin shows the equivalence between kernel self-organizing maps and self-organizing mixture density networks, an extension of SOM for mixture density modeling.

Besides their quantization, clustering and visualization properties when used in an unsupervised manner, SOM also show interesting properties when embedded in a supervised context. Barreto and Souza assess the use of SOM for nonlinear adaptive filtering, while Koga et al. introduce fuzzy inference based heuristic evaluation in the context of the Self-Organizing Relationship network.

Despite the usual simplicity of the algorithms providing self-organization and topographic map formation, their theoretical study remains difficult. A few authors contribute to advances in this domain: Fort summarizes existing results and points out open questions in the mathematical analysis of SOM. Ghosh et al. use a statistical physics approach to analyze LVQ (Learning Vector Quantization) algorithms, and Rynkiewicz studies the equilibrium points of SOM and their relation to the minimization of a distortion measure.

As SOM provide mostly visual outputs and cluster information, assessing their results, including the sensitivity to convergence and initial conditions, and the interpretation of clusters, is a non-trivial task. Rousset et al. study the topology preservation and show how to choose a robust map that is most stable relatively to the choice of the sampling method and of the learning options of the SOM. Lebart assesses SOM via contiguity analysis, and provides a way to visualize the shape of the clusters.

SOM and related methods may be used on rough vector data. In some cases however, rough data are not available directly, or should be preprocessed for a more adequate use of SOM. Conan-Guez et al. study SOM on dissimilarity data, and show how to drastically improve the efficiency of the implementation when faced to voluminous data sets. Aaron presents a graph-based normalization algorithm for nonlinear data analysis, including SOM. Finally, Simon et al. show how to extract adequate regressors from a time series to allow a further meaningful processing by SOM.

One of the main properties of SOM that are not shared by most other data analysis tools is the ability of easy two-dimensional visualization. Several papers cover this important aspect of topological maps. Venna and Kaski present local multidimensional scaling, an extension of curvilinear component analysis that allows the user to choose an adequate compromise between trustworthiness and continuity. Wu and Takatsuka introduce spherical self-organizing maps that use an efficient indexed geodesic data structure to avoid a prohibitive computation time when avoiding the border effects of SOM. Pölzlbauer et al. introduce two new methods for depicting the SOM based on vector fields, to show the clustering structure at various levels of detail. Estevez and Figueroa provide a distance preserving output representation to the neural gas network, and Samsonova et al. provide a set of tools to perform unsupervised SOM cluster analysis.

Finally, the two last papers cover specific uses of SOM: Mahony et al. use SOM to discover DNA-binding motifs, and Olteanu uses SOM to evaluate the number of regimes in a switching autoregressive model.

We would like to thank all members of the SAMOS team at the Université Paris 1 Panthéon Sorbonne, who contributed in an efficient and friendly way to the success of WSOM 2005. We are also grateful to John Taylor, editor of the Neural Networks journal, who agreed to publish this special issue with extended versions of selected papers presented at the conference. We hope this issue can serve as a reference on recent developments in the field of Self-Organizing Maps, and are looking forward to the next WSOM conference that will be organized in 2007 in Bielefeld by Helge Ritter and hos co-workers.

Marie Cottrell
Université Paris 1 Panthéon Sorbonne, SAMOS-MATISSE

Michel Verleysen
Université catholique de Louvain, Machine Learning Group